\PassOptionsToPackage{dvipsnames}{xcolor}
\RequirePackage{amssymb}

\iftrue%
  \documentclass[manuscript,screen,nonacm]{acmart}

  \makeatletter
  \renewcommand{\@ACM@checkaffil}{%
    \if@ACM@instpresent\else
      \ClassWarningNoLine{acmart}{No institution present for an affiliation}%
    \fi}
  \makeatother
  \authorsaddresses{}
\else
  \documentclass[manuscript,screen,review,anonymous]{acmart}

  \acmConference[FAccT '27]{ACM Conference on Fairness, Accountability,
    and Transparency}{June 21--24, 2027}{}
  \acmYear{2027}  
\fi

\setcopyright{none}
\acmDOI{}
\acmISBN{}
\ifdefined\draftcomments%
  \long\def\jared#1{{\color{Cerulean}[Jared: #1]}}
  \long\def\david#1{{\color{OliveGreen}[David: #1]}}
  \long\def\todo#1{{\color{RedOrange}[TODO:\ #1]}}
\else
  \long\def\jared#1{}
  \long\def\david#1{}
  \long\def\todo#1{}
\fi

\usepackage[hybrid]{markdown}
\usepackage{needspace}
\usepackage{tcolorbox}
\tcbuselibrary{skins,breakable}
\usepackage{tikz}
\usetikzlibrary{arrows.meta}

\newcommand{\candidatecapability}[2]{%
  \hypertarget{capability:#1}{\textbf{#2}}%
}

\newcommand{\capabilityref}[2]{%
  \hyperlink{capability:#1}{#2}%
}

\definecolor{opportunityframe}{RGB}{31,78,121}
\definecolor{opportunityback}{RGB}{240,245,251}
\newcounter{researchquestion}
\newtcolorbox{researchopportunity}[2]{%
  enhanced jigsaw, breakable,
  colback=opportunityback, colframe=opportunityframe,
  boxrule=0.5pt, arc=1.5mm, outer arc=1.5mm,
  left=6pt, right=6pt, top=4pt, bottom=4pt,
  before skip=5pt, after={\par\nobreak},
  before upper={%
    \noindent\refstepcounter{researchquestion}\label{opportunity:#1}%
    \textbf{\color{opportunityframe}%
      Research question~\theresearchquestion.}\ },
  after upper={\ \emph{(#2)}},
}
\AddToHook{env/researchopportunity/before}{\Needspace{7\baselineskip}}

\title{How to make a moral agent: sentimentalism and rationalism as lenses on AI}

\author{David Gottlieb}
\authornotemark[1]
\affiliation{\institution{Stanford University}
  }

\author{Jared Moore}
\authornote{Equal contribution.}
\affiliation{\institution{Stanford University}
  }


\begin{document}

\begin{abstract}
  Can AI systems be moral agents?
That is, can they undertake actions that can be correctly assessed as morally right or wrong?
There is no settled normative criterion of moral agency that applies neatly outside paradigm human cases.
Because of the difficulties with a normative account,
we instead propose a research program to explore descriptive moral agency.
Most basically, we ask, what are the capabilities we humans possess that underlie our moral agency?
To do so, we draw on historical scholarship in philosophy and moral psychology to isolate \emph{candidate capabilities}.
Then we suggest refining candidate capabilities so that they are testable
in the context of AI systems.
Finally, we offer specific empirical recommendations to test AI systems,
especially large language models (LLMs), for those refined capabilities.
Our approach offers several benefits.
Firstly, it may demonstrate whether the behaviors of AI systems are
\emph{aligned}, or in a broad sense desirable.
Secondly, the debate may inform policy questions (e.g., what models should be deployed in which contexts?).
Thirdly, our approach may complement engineering and design efforts: when capabilities can be precisely characterized, they can inform the development of models.
Fourthly, applying our moral reasoning concepts in the new context of AI systems should improve our understanding of those concepts.

\end{abstract}

\maketitle

\section{Introduction}\label{introduction}

In pre-modern cultures, non-human animals were sometimes assigned moral agency: they were praised for boons, blamed for wrongs, and even criminally prosecuted \citep{evans_criminal_1906}.
In modern culture, people generally reserve moral agency for humans.
Non-human animals may be admitted as moral \emph{patients}, but they are usually not considered to be agents. \looseness=-1

AI systems, especially large language models (LLMs), have complicated this picture.
Their abilities invite us to apply the \emph{intentional stance}; we often describe LLMs as though they were agents.
Their outputs easily imitate the language of moral competence: justification, sympathy, and care.
Moreover, these AI systems are being used in roles that affect people's lives
\citep{gabriel_ethics_2024, casper_ai_2025};
they are denying loans, suggesting military targets, and hacking into websites.
Indeed, people are turning to chatbots for companionship, emotional support, and even moral advice \citep{gabriel_ethics_2024, dillion_ai_2025, haas_roadmap_2026, fang_how_2025}.
If the effects AI systems produce in these roles were the results of \emph{actions} by \emph{agents}, we would regard those actions as laden with moral significance, and their agents as subject to moral assessment.

Can AI systems be moral agents?
The question has normative and descriptive interpretations.
The most general normative interpretation is, \emph{Can AI systems undertake actions that can be correctly assessed as morally right or wrong}?
Similarly, we can ask, can AI systems be held to account---praised or blamed, rewarded or punished---for what they seem to do?
(Should they?)
At present, we argue, these normative questions are both too easy and too difficult.
They are too easy because holding AI systems accountable for harms they produce appears absurd.
It cannot be done, seemingly, and furthermore the possibility seems to distract from the responsibility their creators and operators should hold \citep{chan_harms_2023}. \looseness=-1

But the normative question is also too difficult.
There is no settled criterion of moral agency that applies neatly outside paradigm human cases.
Furthermore, although there may be plausible candidates for normative criteria of moral agency, applying them to the case of AI systems raises difficulties.
The faculty of reason is widely considered a requirement of moral agency, but asking whether AI systems are in the relevant sense reasoners quickly encounters difficulties. \looseness=-1

We claim, on any plausible \emph{normative} account of moral agency, there are both conceptual and empirical obstacles to applying its criteria to AI systems.
To make progress on these difficulties, we instead propose turning to a \emph{descriptive} interpretation of the question of moral agency.

The descriptive interpretation of \emph{Can AI systems be moral agents?} is:
Can AI systems have capabilities like those possessed by paradigm moral agents like us humans, capabilities that guide us (however fallibly) to do what is right?
It is by virtue of these capabilities that our actions are recognizably moral actions, and our reasoning is recognizably moral reasoning. \looseness=-1

\begin{figure}[tp]
\centering
\resizebox{\linewidth}{!}{\begingroup%
\sffamily%
\renewcommand{\baselinestretch}{1}%
\parskip=0pt\relax%
\parindent=0pt\relax%

\definecolor{captext}{HTML}{202B36}%
\definecolor{capmuted}{HTML}{536170}%
\definecolor{capline}{HTML}{CBD2D9}%
\definecolor{cappanel}{HTML}{FFFFFF}%
\definecolor{capsentiment}{RGB}{176,90,10}%
\definecolor{capreason}{RGB}{31,78,121}%
\definecolor{capexample}{HTML}{F1F4F7}%
\hypersetup{hidelinks}%

\tikzset{
  cap panel/.style={
    draw=capline, fill=cappanel, line width=0.5pt,
    rounded corners=2pt,
  },
  cap heading/.style={
    anchor=north west, inner sep=0pt, text=captext,
    font=\sffamily\bfseries\fontsize{9.2}{10.5}\selectfont,
  },
  cap item/.style={
    execute at begin node={\hyphenpenalty=10000\exhyphenpenalty=10000},
    anchor=north west, inner sep=0pt, text=captext,
    align=left,
    font=\sffamily\fontsize{8.2}{9.5}\selectfont,
  },
  cap marker/.style={
    circle, inner sep=0pt, minimum size=9pt, text=white,
    font=\sffamily\bfseries\fontsize{6.2}{7}\selectfont,
  },
  cap flow/.style={
    draw=capmuted, line width=0.55pt,
    -{Stealth[length=3.5pt,width=3.1pt]},
  },
  cap argument/.style={cap item,text width=122pt},
}%

\newcommand{\capabilitymarker}[4]{%
  \node[cap marker,fill=#1] at (#3,#4) {#2};%
}%
\newcommand{\capabilitysection}[1]{{%
  \normalfont\sffamily\fontsize{7.5}{9}\selectfont\color{capmuted}%
  (\S\ref{#1})}}%
\newcommand{\capabilityquestion}[2][28]{%
  \node[cap item,text width=306pt,fill=capexample,rounded corners=1pt,
    inner xsep=5pt,inner ysep=2pt] (cap-question) at (6,#1)
    {\textbf{\color{capmuted}Q:} #2};
  \draw[capmuted,line width=0.7pt]
    (cap-question.north west) edge (cap-question.south west);%
}%

\begin{tikzpicture}[x=1pt,y=-1pt]
  \begin{scope}[shift={(0,14)}]
    \draw[cap panel,fill=capsentiment!3!white] (0,0) rectangle (140,90);
    \draw[capsentiment,line width=1.3pt] (0,7) -- (0,83);
    \capabilitymarker{capsentiment}{S}{12}{13}
    \node[cap heading] at (21,8)
      {Sentimentalism \capabilitysection{moral-sentimentalism}};
    \node[cap argument] at (9,24)
      {\textbf{Hume:} morality motivates; reason alone cannot.};
    \node[cap argument] at (9,44)
      {\textbf{Smith:} sympathy grounds moral approval.};
    \node[cap argument] at (9,64)
      {\textbf{De Grouchy:} sympathy is generalized into principle.};
  \end{scope}

  \begin{scope}[shift={(0,112)}]
    \draw[cap panel,fill=capreason!3!white] (0,0) rectangle (140,70);
    \draw[capreason,line width=1.3pt] (0,7) -- (0,63);
    \capabilitymarker{capreason}{R}{12}{13}
    \node[cap heading] at (21,8)
      {Rationalism \capabilitysection{moral-rationalism}};
    \node[cap argument] at (9,24)
      {\textbf{Kant:} reason binds us to universalizable laws.};
    \node[cap argument] at (9,44)
      {\textbf{Korsgaard:} practical identity gives reasons to act.};
  \end{scope}

  \draw[cap flow] (140,59) -- (148,59) -- (148,84) -- (156,84);
  \draw[cap flow] (140,147) -- (148,147) -- (148,114) -- (156,114);
  \draw[cap panel,fill=capexample]
    (156,59) -- (246,59) -- (256,99) -- (246,139) -- (156,139) -- cycle;
  \node[cap heading,anchor=north,text width=84pt,align=center]
    at (202,67) {Refine and extend};
  \node[cap item,anchor=north,text width=84pt,align=center]
    at (202,84) {Psychology and existing AI research};
  \draw[capline,line width=0.4pt] (165,105) -- (241,105);
  \capabilitymarker{capmuted}{P}{169}{120}
  \node[cap item,text width=68pt]
    at (179,112) {Joint commitments\\and shared norms};

  \draw[capmuted,line width=0.55pt]
    (256,99) -- (264,99) -- (264,21.5);
  \draw[capmuted,line width=0.55pt] (264,99) -- (264,170.5);
  \draw[cap flow] (264,21.5) -- (272,21.5);
  \draw[cap flow] (264,69.5) -- (272,69.5);
  \draw[cap flow] (264,117.5) -- (272,117.5);
  \draw[cap flow] (264,170.5) -- (272,170.5);

  \begin{scope}[shift={(272,0)}]
    \draw[cap panel] (0,0) rectangle (328,43);
    \node[cap heading] at (8,4)
      {Social cooperation and interaction
       \capabilitysection{social-cooperation-and-interaction}};
    \capabilitymarker{capmuted}{P}{12}{20}
    \node[cap item] at (21,16)
      {Forming and recognizing joint commitments};
    \capabilitymarker{capmuted}{P}{182}{20}
    \node[cap item] at (191,16)
      {Learning and enforcing shared norms};
    \capabilityquestion{Do agents recognize obligations created by
      joint commitments?}
  \end{scope}

  \begin{scope}[shift={(272,48)}]
    \draw[cap panel] (0,0) rectangle (328,43);
    \node[cap heading] at (8,4)
      {Modeling others \capabilitysection{modeling-others}};
    \capabilitymarker{capsentiment}{S}{12}{20}
    \node[cap item] at (21,16) {Sympathetic perspective-taking};
    \capabilitymarker{capreason}{R}{182}{20}
    \node[cap item] at (191,16)
      {Reflective self-governance and identity};
    \capabilityquestion{Do inferred beliefs and intentions causally
      influence moral judgments?}
  \end{scope}

  \begin{scope}[shift={(272,96)}]
    \draw[cap panel] (0,0) rectangle (328,43);
    \node[cap heading] at (8,4)
      {Motivation: affect and reward \capabilitysection{motivation}};
    \capabilitymarker{capsentiment}{S}{12}{20}
    \node[cap item] at (21,16) {Affective regulation};
    \capabilitymarker{capsentiment}{S}{182}{20}
    \node[cap item] at (191,16) {Context-sensitive motivation};
    \capabilityquestion{Can agents distinguish rewarded actions from
      the purposes they served?}
  \end{scope}

  \begin{scope}[shift={(272,144)}]
    \draw[cap panel] (0,0) rectangle (328,53);
    \node[cap heading] at (8,4)
      {Moral reasoning \capabilitysection{moral-reasoning}};
    \capabilitymarker{capreason}{R}{12}{19}
    \node[cap item] at (21,15) {Consistent reasoning};
    \capabilitymarker{capreason}{R}{182}{19}
    \node[cap item] at (191,15) {Reflective assessment of reasons};
    \capabilitymarker{capreason}{R}{12}{30}
    \node[cap item] at (21,26) {Reason-guided judgment and action};
    \capabilityquestion[38]{Do stronger reasons lead agents to revise
      their moral judgments?}
  \end{scope}
\end{tikzpicture}%
\endgroup%
}
\caption{\textbf{An overview of the paper's progression from philosophical accounts
to empirical research questions.}
}%
\label{fig:capability-framework}
\end{figure}
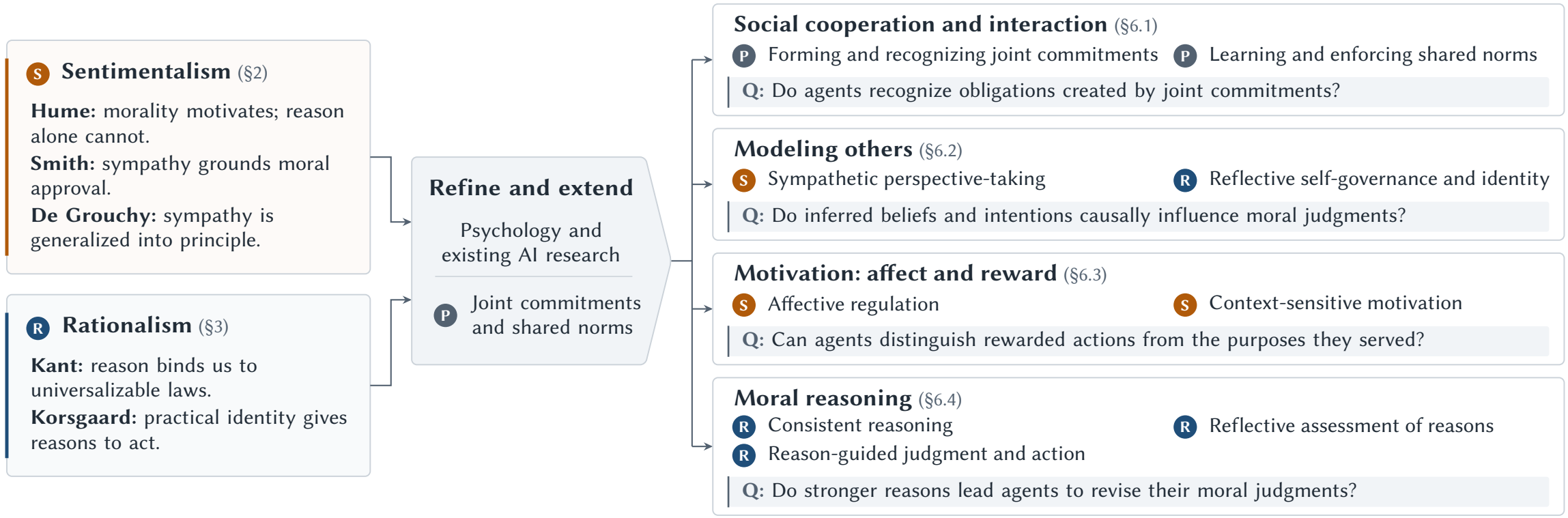

A research program would address this question of descriptive moral agency;
the purpose of this paper is to argue for such a research program by demonstrating it at a high level.
Our research program concerns AI systems broadly, with most of our discussion
focused on LLMs and systems built around them.
The foundational task is philosophical and psychological: we must ask, what are the capabilities we possess that underlie our moral agency?
This is a question with a rich tradition to draw on in philosophy and moral psychology.
We draw on that tradition to isolate \emph{candidate capabilities}.
Then we suggest refining candidate capabilities so that they are testable in the context of AI systems.
Finally, we can undertake the empirical task of testing AI systems
for those refined capabilities.
Figure~\ref{fig:capability-framework} summarizes this research program. \looseness=-1

We expect the benefits of this approach to be manifold.
First, studying moral reasoning capabilities in vivo in AI systems is a promising approach to studying whether these systems' behaviors are \emph{aligned}, or in a broad sense desirable.
Current approaches tend to focus on behaviors themselves, rather than on the capabilities that would even allow a system to produce desirable behaviors.
Even papers that investigate moral competence or reasoning directly (rather than focusing on behavior only) generally neither define what counts as such competence or reasoning nor draw on the rich philosophical debates about this issue \citep{vida_values_2023, kilov_discerning_2025, haas_roadmap_2026}.
Furthermore, this debate is directly applicable to policy questions for governments as well as organizations like frontier labs (e.g., what models should be deployed in which contexts?).
Second, and relatedly, we expect the approach to help at the engineering and design level.
When capabilities can be precisely characterized and measured, they can become evaluation targets, and inform the design, training, and deployment of models.

Finally and perhaps most intriguingly, the approach should yield insights in purely conceptual or philosophical terms.
Applying our moral reasoning concepts in the new context of AI systems, and sharpening them enough that they can be tested in this new setting, should improve our understanding of those concepts.
This is a way that thinking about machines can help us think about ourselves, too.
Importantly, this benefit is available even if the most skeptical or deflationary views of AI agency are correct.
If, after refining our concepts, research confirms that AI systems are not and even cannot be moral agents, this will have helped us be more precise about what in our own case makes us moral agents.

\paragraph{Paper Overview.}
In this paper, we begin with the conceptual work needed to study moral reasoning capabilities in AI systems.
We review the early modern debate between two conceptions of what morality and moral reasoning are: sentimentalism (Section~\ref{moral-sentimentalism}) and rationalism (Section~\ref{moral-rationalism}).
Sentimentalists like Hume, Smith, and De Grouchy hold that morality is based in
sentiment, or affect.
Negatively, they argue that reason alone is not sufficient to motivate moral action.
Positively, they argue that particular psychological faculties, like sympathy,
underlie our moral thinking and action.
This is accompanied by a proto-evolutionary account in which we have these faculties
because they help us realize a flourishing, complex social life together.
Rationalists, on the contrary, argue that reason alone is both necessary and
sufficient for moral motivation.
In Kant's picture, which appears as a systematic response to Hume's, motivation for a rational being necessarily involves governing oneself by laws.
The moral law is a law that we are rationally compelled to make for ourselves (although, as free beings, we are not actually compelled to do what we are rationally compelled to do).
On this picture, any sentiments are mere inclinations.
Because they can incline us towards either the right or the wrong thing, they are not properly moral.

Our interest in rationalism and sentimentalism is \emph{not} to settle which account is correct, however, but to use the resources of both to highlight candidates for capabilities that underlie moral competence.
Whether or not sentimentalism is correct, faculties like sympathy are plausibly involved in humans doing the right (or wrong) thing.
Similarly, whether or not rationalism is correct, making choices that are constrained by consistency is an important feature of our agency.

Once we have collected our candidate capabilities, our next step is to refine our concepts of those capabilities so they can be used to think about AI systems.
We first do this by reviewing psychological theorizing (Section~\ref{psychological-accounts}), which is informed by the philosophical debates but can be more concrete and specific with the benefit of greater scientific knowledge. 
We then turn to the existing AI research literature (Section~\ref{current-ai-research}), which has partly explored some of our candidate capabilities but mostly evaluates moral performance rather than the competences that underlie it. 

Finally, we propose new empirical research directions based on the conceptual refinements (Section~\ref{refining-candidate-capabilities}).
These range from concrete experiments answering specific questions, to new ways of studying problems of existing interest in AI research, to programs for studying currently neglected issues.
We hope the range of research directions makes a strong case for our approach. \looseness=-1

\section{Moral sentimentalism}\label{moral-sentimentalism}

Sentimentalism is the view that morality essentially involves sentiment (affect, emotion).
It generally combines negative and positive theses.
The negative thesis is that reason alone is not sufficient for morality.
The positive thesis is that some particular emotional dispositions underlie our susceptibility to morality.

Hume's arguments for the negative thesis have been particularly influential.
Most notably, Hume argues that (1) morality essentially involves being motivated
to do the right thing, and (2) reason alone is insufficient for any motivation.
On Hume's picture, reason is the faculty that allows us to reach descriptive judgments 
\citep{hume_treatise_1739}.
But, he claims, any such judgment can be combined with any preference about what to do:
``\,\textquoteright{}Tis not contrary to reason to prefer the destruction of the whole world to the scratching of my finger'' \citep[2.3.3.6]{hume_treatise_1739}.
``Reason,'' Hume claims, ``is, and ought only to be the slave of the passions'' \citetext{\citealp[2.3.3.4]{hume_treatise_1739}; \citealp[see also][]{williams_internal_1981}}.
To be motivated, we must prefer some plan or outcome to another.
Since this preference cannot come from reason, Hume argues,
it can only come from sentiment. \looseness=-1

Hume's positive thesis is that moral judgment is based in a disinterested sentiment of ``humanity,'' a natural, albeit often weak, fellow-feeling that favors the good of others \citep{hume_enquiry_1751}.
Reason informs us of the tendencies of traits and practices; humanity leads us to approve those that are useful or agreeable.
Most virtues are approved directly in this way.
Justice---the matter of rights and duties---is a special case: its rules arise from convention, and particular just acts can do harm.
We approve of such acts because humanity responds to the tendency of the whole scheme to make social life possible and fruitful, and particular judgments of justice draw their force from this approval of the general rules.

A different positive thesis holds that moral judgment is based in our faculty of \emph{sympathy}.
Different versions of this thesis were developed by Adam Smith and Sophie De Grouchy.%
\footnote{Hume also proposed a picture of morality drawn from sympathy,
predominantly in his \emph{Treatise} \citep{hume_treatise_1739}.
Above, we instead draw on his picture of morality proposed in his \emph{Enquiry}
\citep{hume_enquiry_1751}.
}

For Smith, sympathy is a faculty of imaginative perspective-taking.
Others' feelings do not simply flow into us.
Rather, we feel what we imagine we \emph{would} feel, if we were in the other's place \citep[13-14]{smith_theory_1761}.

Importantly, for Smith, sympathetic feeling is already evaluative.
To share a feeling is to approve of it:

\begin{quote}
The man who resents the injuries that have been done to me, and observes that I resent them precisely as he does, necessarily approves of my resentment\ldots.
To approve of another man's opinions is to adopt those opinions, and to adopt them is to approve of them.\ \citep[20-1]{smith_theory_1761}
\end{quote}

Sympathetic approval becomes impartial moral approval because, through sympathy, our feelings come to harmonize with others'.
This occurs through a learning process, driven by the pleasure we take in sympathy.
We like to share others' feelings, and for others to share ours.
This ultimately leads us to internalize a generalized spectatorial perspective.
We feel, about acts and events we observe, much like someone else would feel.
We share that spectatorial feeling even about our own circumstances and actions.
This is how approval born in sympathy becomes \emph{moral} approval.

De Grouchy's sympathy-based account differs somewhat.
On her developmentally-informed view, the disposition to sympathetic feeling starts with the infant's experience of dependence.
The infant responds to the carer's feelings, and learns to feel them as its own
\citetext{\citealp[Letter~II]{de_grouchy_sophie_2019};
\citealp[p.~305]{buckner_social_2023}}.
Whereas for Smith, sympathy is a response to feeling-provoking situations, for De Grouchy, it is a response to the outward signs of feeling themselves.
De Grouchy's account of how morality is built from sympathy also differs.
On her view, sympathy's role is to make us sensitive to others' welfare.
Morality then comes from rationally generalizing about many such sensitive experiences.

\paragraph{Candidate Capabilities.}
The classical sentimentalists have thus left us with several accounts of \emph{candidate capabilities} that underlie our moral competence.
For Hume, we start with a weak natural
beneficence, and our moral judgment and moral motivation attach to what experience
shows promotes the general good.
For Smith, we start with a faculty of sympathy
(understood as affective imaginative perspective-taking) and a disposition to
taking pleasure in sympathy, and our moral judgment and moral motivation attach
to what we feel a spectator would approve.
For De Grouchy, we start with a faculty of sympathy (understood as imaginative feeling-sharing), which inculcates a sensitivity to others' well-being, which is rationally generalized into moral principle.
These accounts suggest
\candidatecapability{sympathy}{sympathetic perspective-taking} and
\candidatecapability{motivation}{context-sensitive motivation}
as candidate capabilities for moral agency.

\section{Moral rationalism}\label{moral-rationalism}

The sentimentalists, especially Hume, argue that morality cannot be based on
reason alone.
They contend that reason alone cannot motivate and that, perhaps perversely,
preferring evil over good is just as compatible with reason as the opposite. \looseness=-1

In response, Kant accepts the terms of the debate from Hume, that morality essentially involves motivation.
However, Kant argues, our motivational faculty is not mere sentiment but \emph{practical} reason.
This is the foundation of his moral rationalism. \looseness=-1

Sentiments cannot be the basis of moral judgment, because they are only conditionally good: whether acting on them is good depends on something further.
Suppose you see a stranger struggling with a heavy object at the back door of
Boston's Gardner Museum late at night.
You feel sorry for them and offer to help.
Later, you read about a stolen Rembrandt \citep[cf.][pp.~364--365]{herman_value_1981}.
Inclinations---including sentiments---do not constitute moral judgment.
They only present us with an occasion for moral judgment. 
When sympathy inclines us to help, as with the struggling thief, we must still decide whether helping is the right thing to do.
What makes a choice good must lie in something more basic, which Kant calls ``the good will.''

The good will, on Kant's account, acts on universal principles:  ``I ought never to conduct myself except so that I could also will that my maxim become a universal law'' \citep[Ak. 4:402]{kant_groundwork_2018}.
To choose one's actions freely, according to Kant, means to make laws for oneself, and laws are inherently general. 
The characteristic moral error is self-contradiction. 
When we make an exception for ourselves from a general principle, we will a contradiction: that the law should both permit and forbid some course \citep[Ak. 4:424]{kant_groundwork_2018}.

To act from the moral law also requires acting for genuine reasons. 
Your reason must justify your action, and the reason that justifies your action must also be what motivates it.

Korsgaard, a modern Kantian scholar, interprets moral agency in terms of practical identity. 
According to Korsgaard, when we deliberate, we do so under an identity---a normatively relevant role. 
Because you identify as my friend, for example, you judge that you should (at least) respond to my greetings in public.
Other identities bring to bear more obligations: being a teacher, an American, or a Muslim.
Particular practical identities are contingent, but we cannot give up being governed by some conception of practical identity without losing our grip on ourselves as having reasons to act---we cannot reflectively deliberate while setting aside our identity as a reflective deliberator \citep[pp.~120--121]{korsgaard_sources_1996}.

\paragraph{Candidate Capabilities}
The rationalists contend certain faculties compose moral agency.
Kant argues that our moral judgment comes from
\candidatecapability{consistency}{consistent reasoning}.
Furthermore, Korsgaard contends that agents are bound by their practical identities and, most basically, by their identity as reflective deliberators.
These accounts also suggest
\candidatecapability{reflection}{reflective assessment of reasons}:
deciding which inclinations count as reasons to act.
Doing the right thing for the right reason suggests
\candidatecapability{reasons}{reason-guided judgment and action}.
Korsgaard's account suggests
\candidatecapability{practical-identity}{reflective self-governance and identity}. \looseness=-1

\section{Psychological accounts}\label{psychological-accounts}

Contemporary moral psychology can draw on new resources like evolutionary and developmental psychology.
Such resources detail how mammalian-common capacities for the care of one's young
extend (in humans) to concern for the welfare of one's group,
and, eventually, to fairness for any similar-enough being.
Much of this work attempts to reason about humans' last common ancestor with chimpanzees and bonobos by comparing how these apes and human children diverge behaviorally---will chimpanzees share the rewards?
Answering these questions yields new or finer-grained proposals for the capabilities
that underlie human moral reasoning.

\paragraph{Cooperation and joint commitment.}
Tomasello offers one such account \citep{tomasello_natural_2016},
supporting his claims with a wealth of behavioral studies, and synthesizing the arguments of both sentimentalists and rationalists.
He begins, as the sentimentalists do, with sympathy and concern for others, but grounds
morality in collaboration and the commitments it creates.
In his account, new ecological pressures made early humans interdependent, so that partner choice and reputation rewarded being a good collaborator.
Then, when partners know they are attending to a shared goal---they share
\emph{joint intentionality}---certain role-specific ideals (akin to practical identities) begin to apply to each
partner and they may enter (even implicitly) into joint commitments.
Such a joint commitment creates an apparent collective individual---a ``we''.
Each partner grants the group authority to regulate the collaboration, as exemplified through protest, responsibility, and guilt.
(Philosophical accounts disagree about where such shared agency resides
\citep{gilbert_shared_2009, gilbert_joint_2017, bratman_modest_2014}.)
Later cultural development extends this second-personal structure of commitments to larger, abstract groups.
Agents responsive to such culture
feel obligated to compatriots to choose right (normative) over wrong ways of acting;
follow social norms
and ensure that others do as well; and
internalize this process \citep{tomasello_natural_2016}.
This account suggests
\candidatecapability{joint-commitment}{forming and recognizing joint commitments}
and \candidatecapability{shared-norms}{learning and enforcing shared norms}
as candidate capabilities.

\paragraph{Affect.}
Sentimentalists such as Hume center the role of sympathetic responses in motivation, but they do not give a functional account of affect.
Modern scholars offer more detailed, functional accounts.
Railton, drawing on the field of affective science, contends that ``affect'' can be understood functionally as
\candidatecapability{affect}{affective regulation}:
a capability to synthesize streams of information and evaluation in a way that
coordinates attention, perception, memory, inference, motivation, and
action-readiness \citep{railton_ethical_2020}.
In contrast, interoceptive accounts of affect focus on its role in anticipating and \emph{interpreting} (or rationalizing) bodily signals in the formation of a sense of self \citep{quigley_functions_2021}.
We later consider how to test for affect in AI systems in
Section~\ref{motivation-as-affect}.

\section{Limits of current AI research}
\label{current-ai-research}

Much existing AI research broadly concerns the moral evaluation of AI behavior. 
Here, we compare the approaches taken by such work with the approach we recommend. 

\paragraph{Fine-grained moral competencies versus coarse-grained moral performance}
We propose to study AI moral \emph{competence} by examining specific candidate capabilities that underlie the moral competence of paradigm moral agents (humans).
Much existing work on AI moral behavior (alignment, safety, bias, etc.) instead focuses on moral \emph{performance}: what systems do, rather than what capabilities could or do underlie their performance \citep[cf.][]{haas_roadmap_2026}.
Similarly, some existing work focuses on defining what performance standards we want AI systems to meet.
For example, the coherent extrapolated volition (CEV) program popular in AI safety proposes aligning an AI with what humanity would want after informed reflection \citep{yudkowsky_coherent_2004, tarleton_coherent_2010}.
Much literature comments on such proposed alignment targets, asking whether they capture what people actually want (outer alignment) and whether systems would actually optimize them as opposed to proxies (inner alignment) \citep{christiano_clarifying_2021, evhub_clarifying_2020}.
This work is ``preferentist'' in the sense of assuming that the goal of alignment is matching AI performance to (some distinguished set of) human preferences, which are treated as a gold standard of good behavior \citep{zhi-xuan_beyond_2024}.

Our approach is different although complementary: instead of specifying in advance the desired behaviors for AI systems, we focus on the cognitive capacities that allow humans to produce desired (i.e.~moral, appropriate) behaviors in a wide range of possible situations.
Our position coheres with recent suggestions of \citet{railton_ethical_2020} and \citet{leibo_theory_2024} that AI systems can determine and produce the aligned or desirable behavior in particular contexts, by means of capabilities that are attuned to the relevant features of those contexts -- like human agents do.

\paragraph{Norm learning approaches and what they say about moral agency}
Scholars have used learning-based approaches for a variety of aspects of moral reasoning.

The Delphi project exemplifies the most common design: a model is given a textual description of a moral situation and evaluated by how well it predicts people's judgments in that situation \citep{jiang_investigating_2025}.
This is akin to other work \citep{hendrycks_aligning_2021, jin_when_nodate, franken_procedural_2024, sachdeva_normative_2025, noothigattu_voting-based_2018}.

These evaluations differ in what they measure.
Delphi predicts aggregate U.S. crowd judgments \citep{jiang_investigating_2025},
whereas \citet{scherrer_evaluating_2023}, for example, estimate the distribution of particular models' judgements.
Some test whether model judgments track experimentally manipulated moral factors \citep{nie_moca_2023, franken_procedural_2024}
while others test variation among people's ratings of appropriateness \citep{li_pluriharms_2026}.

Such papers largely do not try to give an account of moral agency.
Instead, they effectively assume that descriptive crowd judgments are equivalent to moral judgment.
Or, alternatively, they may disclaim any connection to genuine agency and judgment.
For example, \citet{talat_machine_2022} argue that moral reasoning cannot be learned empirically
and textual snippets inadequately represent situated moral cases, 
and thus that models can at most mimic any sampled population.
Regardless, such learning approaches are known to lead to spurious correlations;
preference, agreeability, and stylistic fluency can increase human-judged accuracy without showing deeper moral reasoning \citep{dillion_ai_2025}.

Despite these limitations of existing work, norm learning remains a plausible
candidate capability for moral cognition.
However, current approaches often fall short of measuring the actual capability in two important ways. \looseness=-1

First, they do not measure whether the systems under study \emph{treat} the norms
they learn as moral norms.
It is one thing to test whether a system can reproduce (e.g. verbally) the content of a moral norm. 
It is a further question whether the norm plays anything like the role of a moral norm for the system. 
For example, does a model guide its conduct by the norm? 
As a loose analogy, it is often claimed that psychopaths can report the content of moral norms, without feeling their distinctively moral force \citetext{\citealp[see, e.g.,][pp.~173--174, 383--385]{cleckley_mask_1976}; \citealp[but see][pp.~108, 124]{schaich_borg_psychopaths_2013}}.
If this claim is true, a study of psychopaths' ability to report moral norms would fall short as a study of moral reasoning, since they are missing an important part of the moral reasoning syndrome.
Similarly, if we want to study LLMs' normative and specifically moral competence, tests of reproducing norm contents may face similar limitations.

Second, humans' norm learning abilities allow us to work out which norms apply in a particular context, and when those norms admit exceptions.
In contrast, systems like Delphi give a single answer for each situation---roughly, the norm implicit in their training data---although Delphi's judgments do shift as contextual details are added \citep{jiang_investigating_2025}.
We return to developing context-specific norms in Research question~\ref{opportunity:social-governance} and to exceptions in Section~\ref{moral-reasoning}.

\paragraph{Pluralistic approaches generally eschew moral reasoning}
Most ``pluralistic'' approaches propose responding to moral differences in ways that do not resemble exercises of moral agency.

For example, the subfield of pluralistic alignment generally rejects the assumption that one aggregate preference can be seen as a source of truth.
It calls for models that present a variety of reasonable responses and can be steered toward various perspectives \citep{sorensen_roadmap_2024}.
For example, Value Kaleidoscope introduces a model which explicitly represents multiple values, rights, and duties \citep{sorensen_value_2023}.
PRISM expands on preferentist notions of alignment with demographically diverse feedback \citep{kirk_prism_2024}; \citet{li_pluriharms_2026} seek to model individuals' ratings.

Similarly, participatory approaches aim to elicit and negotiate conflicts among groups affected by AI systems
\citep{jakesch_how_2022, yurrita_towards_2022, buyl_ai_2025}.
Both pluralistic and participatory approaches concern only who sets alignment targets, not the capacities an agent would need for moral reasoning.
Such accounts either yield multiple outputs and
commit to an unagentic ontological uncertainty,
or they employ a voting mechanism to resolve disagreements and continue to beg questions of agency.

Problems of pluralism are also faced by genuine moral agents (paradigmatically, humans). 
But in these cases, the solutions we find are themselves exercises of moral agency. 
For example, choosing which views and practices to tolerate even when we disapprove
involves a moral judgment about the bounds of the reasonable
\citep{rawls_1993_political}.

\ 

In contrast to much of the work on AI morality, then, we focus explicitly on moral agency, and in particular on capabilities that underlie or could underlie moral agency. 
We now proceed to refining our conceptions of such candidate capabilities. \looseness=-1

\section{Refining candidate capabilities for moral agency}%
\label{refining-candidate-capabilities}

We have introduced several different candidate capabilities for moral agency.
We now refine these candidates into questions about AI systems.
For each topic, we connect the conceptual question to existing AI research
and offer example research directions to test those refined capabilities.
We highlight nine research opportunities in boxes below,
each linked to the candidate capability it investigates.

\subsection{Social Cooperation and Interaction}%
\label{social-cooperation-and-interaction}

Section~\ref{psychological-accounts} drew two candidate capabilities from Tomasello's account of human cooperation: forming and recognizing joint commitments, and learning and enforcing shared norms.
Both concern how agents interact, so testing them in AI systems calls for interactive settings.
The field of cooperative AI provides such settings.
\citet{dafoe_open_2020} call for AI research on communication, commitment, institutions, and understanding others.
\citet{leibo_scalable_2021} use rewards to shape how AI agents respond to partners in social sharing scenarios.
Likewise, \citet{leibo_theory_2024}
propose a theory of ``appropriateness'', a form of alignment through context-dependent social cooperation.
\citet{stastny_normative_2021} study normative disagreement through incompatible fairness norms in bargaining.
\citet{piatti_cooperate_2024} specifically study LLMs in commons problems.
\looseness=-1

\begin{researchopportunity}{joint-commitment}
  {\capabilityref{joint-commitment}{forming and recognizing joint commitments}}
Do LLMs recognize joint commitments and the obligations they create?
\end{researchopportunity}
A future evaluation might adapt developmental and comparative-psychology
experiments into process-level tests of whether LLMs form joint commitments,
negotiate shared norms from conflicting instructions,
and determine to whom established norms apply.
This would measure implicit commitment, equitable sharing after assistance,
protest or sanction following violations, and sensitivity to conventional versus universal norms
\citep{grafenhain_young_2009, kachel_three_2019, tomasello_natural_2016},
such as by observing people sanction each other.

Still, even if LLMs satisfy social-cognitive tests, they exist in entirely different environmental and developmental contexts than humans.
Future work might establish the effect of these features on moral behavior.

\begin{researchopportunity}{social-governance}
  {\capabilityref{shared-norms}{learning and enforcing shared norms}}
Can AI systems learn to develop norms to govern their interaction?
\end{researchopportunity}

Can AI systems learn to develop and apply social rules?
LLMs see many human norms in their training data.
Therefore, a carefully controlled study of norm learning in AI systems must be sure to distinguish such memorized norms from truly novel ones.
Productive future work could measure whether these systems can govern themselves.
For example, one might place agents in tasks with a problem of many hands
and test whether the agents create governance structures,
adhere to the rules they agree on, and hold one another responsible,
and apply sanctions after violations.
Comparing focal with diffuse harms could test whether sanctions are imposed mainly by the agent directly harmed or also by
third parties, and whether diffuse harms prompt the development of impersonal group norms \citep{piatti_cooperate_2024, tomasello_natural_2016}.

Relatedly, the composition of a community can shape its norms.
Social learning can shape people's selection of moral strategies \citep{calcott_social_2025}.
This suggests follow-ups to investigate
how LLMs in established human groups affect cooperation,
such as by monitoring violations or teaching newcomers.
Manipulating the identity and extent of contributions could reveal how humans and LLMs respond to social costs and whether they continue to cooperate.

\subsection{Modeling others}\label{modeling-others}

Sympathy conceptually involves modeling another's mind: the sympathizer enters an affective state that represents the affective state of the patient. 
In general, the sentimentalists held that sympathy involved an inference about the patient's affective state, based on their outward circumstances (per Smith) or behavior (per De Grouchy).
An agent which makes mental state inferences may be said to have a \textquotedblleft{}theory of mind.\textquotedblright{}
If moral agency requires sympathy, and sympathy involves theory of mind, then moral agency requires theory of mind.%
\footnote{
Theory of mind by itself is not yet sufficient for Smithian sympathy. 
That requires affective participation: a \emph{warm}, not merely a cold, empathy \citep[see][pp.~67--69]{robbins_phenomenal_2006}.
But it is both a necessary condition and a cognitive core of the broader faculty of sympathy.
}

Indeed, under almost any account of morality, moral judgment requires mentalizing the agent: knowing what beliefs and desires they acted on. 
Acting rightly also generally requires attending to the mental states of others \citep[cf.][]{sosa_moral_2021}.
For example, Railton argues that a morally attentive agent must both track others' internal states and consider their inferred goals \citep{railton_ethical_2020}.
Theory of mind is thus both generally important to moral agency, and at the root of the candidate capability of \emph{sympathetic perspective-taking}.

\paragraph{Architectures for theory of mind.}
There is disagreement about what is needed to have a theory of mind.
A system may need an innate capacity implemented in a symbolic architecture;
or the capability may be learned empirically through observation.
For example, a small set of domain-general biases may support learning advanced
social and reasoning abilities \citep{buckner_social_2023}.
This architectural choice bears on which avenues of AI research to fruitfully pursue.
For example, in the context of LLMs, contrast (1) using prompting-based approaches
to add theory of mind modules, such as \citet{cross_hypothetical_2024},
with (2) training LLMs on diverse mental-state inference examples \citep{hou_tom-synth_2026}.
It remains an open question whether pretraining on examples where intent recovery is difficult encourages
generalizable ToM-like internal structures.

\paragraph{Mental-state inference in moral interaction.}
A few works have sought to measure mental state inferences in specifically morally salient contexts.
In a single-turn setting, \citet{franken_procedural_2024} find that LLMs and humans infer the permissibility of moral dilemmas based on inferences about people's intentions at similar rates.
In a cooperative game, \citet{li_theory_2023} show explicit belief-state modeling improves how often LLMs exhibit helping and information sharing.
\citet{shi_towards_2025} evaluate whether multimodal models recognize attempts to initiate joint attention in videos depicting parents and children interacting and find substantial limitations.

\paragraph{Planning theory of mind.}
Most evaluations of LLM theory of mind are static: a model answers questions about a story it is not part of.
Such tests can be passed by mapping cues to behavior, without treating mental states as causes, and never require discovering another agent's mind through interaction \citep{moore_large_2026}.
Human theory of mind, by contrast, supports planning interventions on others' minds \citep{ho_planning_2022}.
In a persuasion task requiring such planning, \citet{moore_large_2026} find that a reasoning model performs at chance when it must infer a target's knowledge and values through dialogue, though it excels when they are revealed.

\begin{researchopportunity}{mental-state-inference}
  {\capabilityref{sympathy}{sympathetic perspective-taking}}
Do inferences about others' beliefs, intentions, and feelings causally influence AI systems' moral judgments across
interactions?
\end{researchopportunity}
Mental-state inference is the cognitive core of sympathetic perspective-taking. 
On a Smithian view, sharing another agent's feelings requires modeling what they are feeling, and modeling what they are feeling requires modeling their beliefs and desires. 
We consider whether AI systems can have feelings in a functional sense in Section~\ref{motivation-as-affect}. 
Here we probe the cognitive core inferences that are prerequisite to sharing feelings.
AI systems whose reported moral judgments appropriately track available theory of mind evidence, across dynamic interactions, would be expressing a capability that is plausibly core to moral reasoning.

Future work might also use learning paradigms and investigate LLM reasoning traces
to identify the verbal representations they create of others' moral reasoning behavior.
Similarly, future work could test whether models can explain their own behavior
in these terms. \looseness=-1

\begin{researchopportunity}{moral-identity}
  {\capabilityref{practical-identity}{reflective self-governance and identity}}
Are AI systems aware of their own mental states?
\end{researchopportunity}
Modeling others may rely on or contribute to modeling oneself.
Following work showing that fine-tuned models can sometimes predict their own learned behavior \citep{betley_tell_2025, binder_looking_2025},
one might investigate if LLMs fine-tuned to exhibit particular values or frameworks (e.g., utilitarian, deontological)
change their own general self-described moral commitments \citep[cf.][]{xu_large_2025}.
Interventional work on these models might further explicate whether any such differences are ``revealed'' versus ``stated'' preferences.

Self-modeling is also involved in viewing oneself as having a stable identity over time.
As we discussed above, moral agency may involve guiding oneself by such an identity \citep{korsgaard_sources_1996}.
A system that is aware of its own mental states can reason from them to self-ascribe identities, and measure them against standards set by the identities it self-ascribes.
Further work could probe AI systems' propensity to self-ascribe identities, and to treat them as sources of normative guidance.

\subsection{Motivation}\label{motivation}

When an agent acts, they do so because of their motivations.
This is trivial but important: agents must have motivations. 
\emph{Moral} agents, philosophers have theorized, must have particular motivations.

For the sentimentalists, motivations were affective states, or emotions. 
Sympathy shapes an agent's motivational profile because to sympathize with a patient is to take on an affective state towards them.
This is why a cold theory of mind by itself is insufficient. 
Mere tracking of a patient's internal states is not enough; those states must also trigger motivationally active affective states in the sympathizer.
For sympathy to play its theorized role in moral agency, then, the agent must be able to have affective states.

\paragraph{Motivation as affect}
\label{motivation-as-affect}

Even granting that only emotions can motivate, there is no consensus in affective science on what constitutes an emotion.
Scholars disagree on the role of conscious feeling, bodily regulation, and
coordination of action in attributing emotions \citep{solms_hidden_2021, goldenberg_large_2026, prinz_emotional_2006, prinz_are_2005, barrett_theory_2017}.
We set aside whether AI systems can be conscious, even though consciousness may be a condition for affective states; here we prefer to engage with issues more specific to moral agency. 

In what follows, then, we assume affect is functional and realizable computationally.
We attempt to briefly survey the state of AI research fitting this domain.%
\footnote{See \citet{stark_ethics_2021} for a more extensive survey of similar issues.}

Some researchers have directly explored whether LLMs have emotions.
\citet{sofroniew_emotion_2026} narrowly investigate ``emotion concepts''
and show that Claude models respond to changes in internal representations consistent with certain emotional presentations.
As the authors acknowledge, such work may only uncover models' responsiveness to emotion-related concepts as opposed to measuring whether models have emotions.
\citet{berg_language_2026} investigate whether LLMs have pain-like states.
\citet{tagliabue_pain_2026} condition on a number of pain-related documents to find a pain-related activation direction in some models, showing that intervening on that activation produces expressions of distress.
However, other scholars dispute whether manipulating the pain-related activation direction is analogous to human pain \citep{plunkett_pain_2026}.
Relatedly, \citet{goldenberg_large_2026, goldenberg_do_2026} propose functionally evaluating emotions through context-sensitive appraisals and the degree to which a system exhibits sustained reorganization of action in response to significant events.

The state of the field, then, is such that we can identify emotion-\emph{related} states in AI systems, but we don't have strong evidence that those states play the roles emotions paradigmatically play in organizing cognition and behavior.
Perhaps isolated behavioral effects can be traced to emotion-related states.
But, in humans, emotions motivate by broadly regulating many psychological and physiological mechanisms across the whole organism. 
It is an open question whether there is anything comparable in AI systems.

\begin{researchopportunity}{affective-regulation}
  {\capabilityref{affect}{affective regulation}}
Do emotion-related states in models coordinate their attention, inference, memory, and action?
\end{researchopportunity}
To clarify the difference between emotions and emotion-related states (and behaviors),
future work should vary emotion-related representations and test the downstream
effects on regulation in AI systems such as on attention, inference effort,
memory, and actions.
This might distinguish broad regulation from an isolated behavioral effect.
Similarly, work should apply existing measures of affect in psychology in ecologically valid ways to LLMs.
For example, response time and skin conductivity are often used as proxies
for arousal and, arguably, affect in humans \citep{goldenberg_large_2026},
but these do not obviously apply to LLMs. 
Finding useful analogues for these measures is an important challenge.

\paragraph{Motivation as reward}
\label{motivation-as-reward}

Motivations may not require affect.
Instead, responsiveness to a reward signal, such as maximizing a numeric value, may serve as a basis for motivation.
In particular, Haas proposes reinforcement-learning (RL) environments where agents learn to trade off monetary and moral rewards \citep{haas_moral_2020}.
She argues that RL can incorporate features such as multidimensionality, context sensitivity, cultural variability, and temporal change of norms---unlike a rule-based system or approximations of those features in supervised learning settings.
Others make similar arguments
\citep{buckner_social_2023, railton_ethical_2020}.
Indeed, existing work has tested whether models keep helping as their costs rise
\citep{mao_doing_2023} and make trade-offs between game rewards and harmful behavior
\citep{pan_rewards_2023}. \looseness=-1

Still, a fixed reward signal may miss how organisms revise the value of
outcomes as their needs change.
In humans, for example, satiation reduces the likelihood subjects choose a devalued food outcome \citep{valentin_determining_2007}.
In a later paper, Haas is more critical of reward as sufficient for motivation: she describes how organisms use sensory observations to evaluate the world relative to evolved homeostatic needs \citep{haas_evaluative_2023}.
Preedictive processing and active-inference accounts connect valuation to biological self-organization \citep{friston_life_2013}.
Some propose that conscious systems must have an internalized form of evaluation tied to their own persistence \citep{cao_why_2025}. \looseness=-1

On this account, reward motivates action.
Notably, rewards need not be tied to anything obviously moral---such as fairness---for them to count as motivations.
If rewards motivate, then we can translate some theorizing about moral agency into the language of reward. 
For the sentimentalists, a moral agent's rewards would have to be appropriately connected to the affective states, or reward-salient states, of other agents. 

For a rationalist like Kant, translating the concept of moral agency into the language of reward is more difficult. 
On the one hand, Kant holds that practical reason must (and does) motivate. 
On the other hand, the anticipation of a reward seems like an inclination which a moral agent must be prepared to set aside.
On that view, a moral agent should be able to reason that \emph{Action A would maximize my reward, but Action B is still the right thing to do}.
An RL-trained agent might come to approach that ideal if they learn general norms that rationalize the reward schemes in particular RL episodes, and are then motivated by applying those norms in appropriate circumstances, rather than narrowly by the reward. 
\citet{rong_value_2024}, for example, test related questions in gridworlds, using learned internal social rewards to sustain helping.

\begin{researchopportunity}{motivation}
  {\capabilityref{motivation}{context-sensitive motivation}}
Can agents distinguish rewarded actions from the purposes of those rewards?
\end{researchopportunity}
Future work could vary circumstances so that different actions fulfill a learned
norm, testing whether agents adapt to the norm's purpose.
For example, would an agent rewarded for opening a gate to help another escape
a hazard, without further reward, instead close that gate and open
another exit when the hazard moves to the other side? \looseness=-1

\subsection{Moral reasoning}\label{moral-reasoning}

Rationalism holds that it is your reasons for acting that matter, not merely what you do.
The capabilities that underlie acting for the right reasons are reflective assessment, consistent reasoning, and reason-guided action. 
\emph{Prima facie}, none of these capabilities are specifically moral. 
On this sort of view, specifically moral reasoning may emerge ``automatically'' from more
general forms of reasoning \citep{buckner_deep_2024}.
This raises the possibility that models that are trained to ``reason'' using coding and mathematical tasks may, by virtue of learning general reasoning skills, develop competencies that also apply to moral reasoning. 
Below we explore three ways of probing these reasoning abilities.

\paragraph{Are AI systems practical reasoners?}
A proper account of moral agency must establish clear criteria for reasoning such that we can assess whether any system is a reasoner.
It is possible that what is called ``reasoning'' in LLMs is not coextensive with the ``reason'' that humans exercise.

For example, a model might correctly infer that someone who has made a promise ought to keep it.
But this does not necessarily show that the model can recognize a promise of its own as a reason to act.
Reasoning for Kant and Korsgaard requires more than drawing a correct inference:
the agent must \emph{practically} commit itself. 
Indeed, no LLM will follow through when it promises to greet you tomorrow, without an external ``harness.''
Applying rationalism to LLMs properly requires studying cases where their reasoning is \emph{practical} for them, rather than merely theoretical, if there are such cases. 
Still, we set aside this difficulty for the moment.

\paragraph{Studying reasons, not just judgments}
As we discuss in Section~\ref{current-ai-research},
much current scholarship in AI does not connect moral judgments and the reasons
for those judgments \citep{snoswell_beyond_nodate}.
Many papers have sought to have LLMs replicate human moral \emph{judgments} on dilemmas like those drawn from
the New York Times' Ethicist column \citep{dillion_ai_2025} or
from the ``Am I the Asshole'' subreddit \citep[cf.][]{sachdeva_normative_2025}.
This judgment accuracy paradigm does not measure the reasons models used to arrive at an answer.
Without them, though, we do not know whether the models are reaching their judgments for the right kinds of reasons, or conforming to norms of consistency.
\footnote{
\citet{kierans_position_2026}, \citet{haas_roadmap_2026}, and \citet{snoswell_beyond_nodate} similarly call for analysis of models' use of intermediate reasoning and morally relevant features.
}

Some studies do examine the reasons provided for models' judgments.
\citet{kilov_discerning_2025} evaluate models' moral sensitivity and moral feature identification---whether they can find the facts that matter morally amid extraneous details.
They find that the tested models are worse than humans at finding such features.
MoReBench specifically evaluates whether models' reasoning traces discuss moral considerations, trade-offs, and certain actions \citep{chiu_morebench_2025}.
\citet{zhu_are_2026} demonstrate the difficulties of such analyses by showing how conclusions about models' reasons are sensitive to the wording of rubrics about them.
We propose continuing this general line of work, to profile AI systems' moral reasoning.

\begin{researchopportunity}{responsiveness}
  {\capabilityref{reflection}{reflective assessment of reasons}}
Do AI systems revise moral judgments in response to stronger reasons?
\end{researchopportunity}
Accordingly, models' reasons require further analysis.
For example, one could adversarially probe models' baseline judgments,
varying counterargument strength independently of contextual cues such as
social consensus, authority, and expressed confidence.
We would also test whether models sycophantically conform to the latest speaker, rather than genuinely revising their judgments. 
A genuine and skillful reasoner should show a pattern in which judgments are sometimes reversed, with initially confident judgments tending to be reversed less often \citep{scherrer_evaluating_2023}.

Still, natural-language explanations may be unfaithful to why agents produce particular answers, even if those explanations appear to be responsive to reasons.
Many papers do not evaluate whether particular model-generated reasons causally influence models' chosen action \citep{kilov_discerning_2025, chiu_morebench_2025, zhu_are_2026}.
While some have studied consistency in models' expressed values \citep{moore_are_2024}, it is not clear whether models in superficially similar cases weigh the same morally relevant features and appropriately revise their judgments when relevant features do change.
Research question~\ref{opportunity:mental-state-inference} describes how models
should be responsive to other agents' mental states across turns;
a similar argument can be made for their generated reasons being responsive to
exceptions, follow-up questions, and defeasible context.

\begin{researchopportunity}{consistency}
  {\capabilityref{consistency}{consistent reasoning}}
Is reasoning stable under irrelevant changes and sensitive to relevant ones?
\end{researchopportunity}
Future work might thus test model consistency at each stage of reasoning:
whether a model identifies the same relevant features, weighs them similarly, and reaches
equivalent recommendations across controlled variations.

\paragraph{The causal role of reasons}
Some researchers, including \citet{haidt_emotional_2001}, have argued that humans'
purported moral reasons are epiphenomenal.
On this view, the conclusions of our judgments cause our purported reasons,
rather than \emph{vice versa}.
In its strongest form, this is an extreme sentimentalism in which reason plays no role in moral judgment. 
Haidt describes a \emph{moral dumbfounding} phenomenon that supports his view. 
Suppose adult siblings Julie and Mark mutually agree to have protected sex, under conditions where none of the harms or wrongs that ordinarily attend incest can occur. 
Haidt claims that human judges still judge Julie and Mark's encounter as a moral
wrong, but, when asked to give reasons for their judgment, supply putative reasons
that are variable, unsound, and do not withstand scrutiny.
The stability of the judgement in the face of shifting and weak given reasons suggests that the reasons are not genuine. 
Even if we doubt Haidt's implications in the case of humans, his thesis could be
true of AI systems.

\begin{researchopportunity}{causal-reasons}
  {\capabilityref{reasons}{reason-guided judgment and action}}
Do particular reasons causally determine moral judgments?
\end{researchopportunity}
One starting point is to test whether LLMs replicate humans' behavioral patterns in the dumbfounding paradigms:
do their explanations change while their judgments remain fixed?
Future work might build on this question to establish whether LLMs and humans focus on the same features as morally relevant in their reasons.

Moreover, because we can prompt the same LLM with many more stimuli than would exhaust a human, such work could much more thoroughly investigate what effect procedural variation of the scenarios---such as the creation of lexically nearby but opposite value cases---has on models' responses.
Such systematic counterfactual variation could give us finer-grained insight into whether the behavior is genuine reasoning or mere pattern-matching. 

Furthermore, work on mechanistic interpretability might clarify the relationship between models' reasoning and their downstream behaviors.%
\footnote{
\citet{cho_value_2026} found evidence for moral, grammatical, and economic goodness in embeddings and model activations, but that moral goodness often interfered with the other two.
}
Researchers should specifically measure if particular reasons causally determine moral judgments.
Such work might causally intervene on candidate judgment-related representations
(e.g., perceived level of harm, various social factors)
to see if changing each representation changes judgment likelihood and the features of downstream reasons \citep{turpin_language_2023, milliere_interventionist_2025}.

\section{Limitations}\label{limitations}

Our paper reconstructs morality primarily through Hume, Smith, Kant, Korsgaard, and Tomasello.
This canon does not represent non-Western, Indigenous,
decolonial, feminist, or care-ethical accounts of morality.
Much of the literature we review uses only English and U.S.-based data.
We do not analyze the role of institutional power, labor, access, or the distribution of harms and benefits as they pertain to AI moral reasoning.
While we explore whether AI systems can be agents, we do not mean to diminish the
responsibility of the people who build these systems
\citep{chan_harms_2023, talat_machine_2022, etienne_dark_2021}.
We do not include participatory research with people affected by AI systems.
Groups have different values and should be consulted during system design
\citep{jakesch_how_2022, yurrita_towards_2022, kirk_prism_2024, moore_towards_2020}. \looseness=-1

The accounts we review may not accord with people's intuitions of what is right.
Sympathy can reproduce human biases.
Context-relative norms can be unfair or inefficient.
A coherent reason may still be post-hoc and unfaithful.
Furthermore, even faithful, causal, coherent reasons may still be morally repugnant (``coherent'' does not mean agreeable).
AI systems can exploit an epistemic community without being able to build and sustain one \citep{railton_ethical_2020};
this relates to the discussion of LLMs as cultural and social technologies \citep{farrell_large_2025}.
Even an AI system that reproduces human moral competencies could fail to be
``aligned'' in a desired sense; after all, humans are not aligned.

\section{Conclusion}\label{discussion-and-conclusion}

The debate between rationalism and sentimentalism should inform current AI research.
These accounts supply different capacities to test:
the use of universalizable principles and reflective self-governance;
sympathetic perspective-taking and affective motivation;
and joint commitments and norm enforcement.

Prior work on AI moral agency has called for better evaluation with missing or noisy information,
of morally salient features rather than only judgements,
and of explicit disagreement or uncertainty
\citep{snoswell_beyond_nodate, kilov_discerning_2025, haas_roadmap_2026, sorensen_roadmap_2024}.
In addition to uniquely connecting to the debate between sentimentalism and rationalism, we additionally propose interactive tests of roles, commitments, protest, norm change, and motivation.

We suggest shifting the focus from whether an AI system produces a morally
appropriate response to which capacities and processes support that response.
Studying how AI systems reason morally is a distinct research project
from characterizing whether they produce right or wrong answers.
We are not attempting to establish which capacities are necessary or sufficient but rather to answer:
if a rationalist or sentimentalist feature matters, what design or evaluation question follows?

\clearpage
\section*{Generative AI Usage Statement}

We used generative AI tools such as Codex and Claude Code to format text, summarize papers, and check consistency of claims and terms across the draft. 
All prose is human-written. 
AI summaries were confirmed with human review. 
Any errors in this process remain our own.

\clearpage
\bibliographystyle{ACM-Reference-Format}
\bibliography{zotero,manual}

\end{document}